\documentclass[sigconf]{acmart}

\usepackage{microtype}
\usepackage{graphicx}
\usepackage{booktabs} 

\usepackage{hyperref}
\usepackage{wrapfig}
\usepackage{url}            
\usepackage{amsmath,amsfonts}       
\usepackage{nicefrac}       
\usepackage{microtype}      
\usepackage{xcolor} 
\usepackage{tabularx}        
\usepackage{multirow}
\usepackage{longtable,booktabs}
\usepackage{algorithm}
\usepackage{algorithmic}
\usepackage{etoolbox}
\usepackage{caption}
\usepackage{subcaption}

\copyrightyear{2022}
\acmYear{2022}
\setcopyright{acmcopyright}\acmConference[KDD '22]{Proceedings of the 28th ACM SIGKDD Conference on Knowledge Discovery and Data Mining}{August 14--18, 2022}{Washington, DC, USA}
\acmBooktitle{Proceedings of the 28th ACM SIGKDD Conference on Knowledge Discovery and Data Mining (KDD '22), August 14--18, 2022, Washington, DC, USA}
\acmPrice{15.00}
\acmDOI{10.1145/3534678.3539174}
\acmISBN{978-1-4503-9385-0/22/08}
\begin{document}

\title{ILASR: Privacy-Preserving Incremental Learning for Automatic Speech Recognition at Production Scale}

\newcommand{\tsc}[1]{\textsuperscript{#1}}
\author[Gopinath Chennupati et al.]{Gopinath Chennupati\\
Milind Rao\\
Gurpreet Chadha\\
Aaron Eakin
}
 \affiliation{%
   \institution{Amazon Alexa}
   \country{USA}
}

\author{Anirudh Raju\\
Gautam Tiwari\\
Anit Kumar Sahu\\
Ariya Rastrow\\
Jasha Droppo}
 \affiliation{%
   \institution{Amazon Alexa}
   \country{USA}
}

\author{Andy Oberlin\\
Buddha Nandanoor\\
Prahalad Venkataramanan\\
Zheng Wu\\
Pankaj Sitpure}
 \affiliation{%
   \institution{Amazon Alexa}
   \country{USA}
}

\renewcommand{\shortauthors}{Gopinath Chennupati et al.}

\begin{abstract}
Incremental learning is one paradigm to enable model building and updating at scale with streaming data. For end-to-end automatic speech recognition (ASR) tasks, the absence of human annotated labels along with the need for privacy preserving policies for model building makes it a daunting challenge. Motivated by these challenges, in this paper we use a cloud based framework for production systems to demonstrate insights from privacy preserving incremental learning for automatic speech recognition (ILASR). By privacy preserving, we mean, usage of ephemeral data which are not human annotated. This system is a step forward for production level ASR models for incremental/continual learning that offers near real-time test-bed for experimentation in the cloud for end-to-end ASR, while adhering to privacy-preserving policies. We show that the proposed system can improve the production models significantly ($3\%$) over a new time period of six months even in the absence of human annotated labels with varying levels of weak supervision and large batch sizes in incremental learning. This improvement is $20\%$ over test sets with new words and phrases in the new time period. We demonstrate the effectiveness of model building in a privacy-preserving incremental fashion for ASR while further exploring the utility of having an effective teacher model and use of large batch sizes.
\end{abstract}


\maketitle
\section{Introduction}
\label{introduction}
Privacy preserving machine learning~\cite{al2019privacy} has been at forefront, due to both increased interest in privacy and the potential susceptibility of deep neural networks to leaks and attacks. Federated Learning (FL)~\cite{mcmahan2017communication} is a machine learning technique that involves training models on edge devices, where data need not leave the device, and can be heterogeneous and non-identically and independently distributed (non-IID). In FL, multiple model updates from a number of participating devices are aggregated. In spite of raw data not leaving the edge device, FL has found to be susceptible to gradient inversion attacks~\cite{zhu2019deep,zhao2020idlg}. In response, various privacy-preserving mechanisms such as differential privacy and secure aggregation~\cite{geyer2017differentially,triastcyn2020federated} have been proposed to counter data leakage and conform to privacy preserving mechanisms. Moreover, the lack of labels for the data present in the participating entities, makes FL more challenging for applications such as automatic speech recognition (ASR). Most research in FL until now focuses on training models from scratch. In this work, we focus on privacy-preserving incremental learning (IL), in the context of end-to-end production model building at scale over extended time periods. Incremental learning \cite{castro2018end,wu2019large} has been extensively used to incrementally update models on the fly instead of training them from scratch. Incremental learning as such is not privacy-preserving.

\noindent Despite the above advances, to the best of our knowledge, few frameworks exist for privacy-preserving incremental training of end-to-end automatic speech recognition models. Prior work on federated learning for speech-based tasks~\cite{dimitriadis2020federated, abs-2102-04429, granqvist2020improving} and end-to-end ASR  ~\cite{gao2021endtoend, guliani2021training}, focus on standard benchmarks\footnote{e.g. LibriSpeech \cite{panayotov2015librispeech} is a small sized dataset ($\sim1000$ hours) recorded in a controlled environment} and not on large scale production data. Privacy-preserving IL on device for end-to-end ASR poses a number of challenges. Production-sized end-to-end ASR systems~\cite{graves2012sequence,chiu2018state} are expensive to train even in traditional distributed setup, on-device training needs more work \cite{cai2020tinytl} to accommodate restrictive memory and computational constraints. Generating training labels i.e. speech transcripts, in near real-time, on the devices is another challenge. To alleviate unavailability of near real-time speech transcripts, teacher transcripts can be used in a semi-supervised and/or self-learning fashion. For example, consider the problem of improving models deployed in edge devices that run voice assistants. In such cases, the number of devices is in the millions, which results in a large scale of streaming data being generated. We propose to use large batch processing for the utterances being collected at the edge devices and sent to the cloud for processing, and the data is only stored ephemerally. However, deploying all or part of the above components on resource constrained speech devices (such as Alexa, Google Assistant and others) is challenging.\\

\noindent We build and use a cloud-based system, \underline{I}ncremental \underline{L}earning for \underline{A}utomatic \underline{S}peech \underline{R}ecognition (ILASR) to train and update production ready ASR models. ILASR automates the entire pipeline of incremental learning in a privacy-preserving manner. To enforce privacy-preserving aspects in the context of ASR, we enforce labelling of the utterances through pre-trained teacher models with no human annotations. ILASR processes an utterance once before updating the model, preserving the chronological order of data. To that end, the contributions of the paper are:
\begin{itemize}
	\item A novel cloud-based IL system to train production ready ASR models in near real-time, with a large amount of streaming de-identified data, without having to manually transcribe or persist the audio.
	\item We provide new insights in terms of usage of large batch processing in ILASR that it does not have detrimental impact on test accuracy as compared to the contradicting findings~\cite{keskar2016large,LI:KDD,shallue2018measuring,mccandlish2018empirical,golmant2018computational,masters2018revisiting} (on CNN ImageNet). We could accommodate fixed learning rates and minimal hyper-parameter optimization~\cite{khodak2020weight} along with large batch training. With a monthly frequency of incremental model updates, we observe that the production models (converged on old data) improve in near real-time on new data belonging to a period of six months
	\item We empirically establish over six months of data that chronological vs randomized order of processing utterances does not produce any observable difference in performance. 
\end{itemize}

\noindent We evaluate ILASR on three student recurrent neural network transducer (RNN-T)~\cite{graves2012sequence} architectures. The semi-supervised learning (SSL) approach produces machine transcripts using a larger teacher ASR model. The students are pre-trained on in-house de-identified data until $2020$. Through training in ILASR, we observe an improvement of $3 - 7\%$ in word error rate (WER) over the pre-trained baselines when these students are trained incrementally on a new time period of six months in $2021$. The improvement in WER is termed relative word error rate reduction (WERR). This increases to $20\%$ on test sets with new words and phrases in $2021$. Similarly, when the student models are trained incrementally each month, we observe WER improvements, as well the phenomenon where models get stale without further updates. 

The paper is organized as follows: section~\ref{sec:background} describes the essential concepts used in the paper; section~\ref{sec:flc-training} explains the proposed system; section~\ref{sec:experiments} describes the experimental settings; section~\ref{sec:results-discussion} presents the results; section~\ref{sec:related-work} summarizes the related literature and finally, section~\ref{sec:conclusions} concludes and recommends future directions.
\section{Background}
\label{sec:background}
In this section we summarize the RNN-T architecture and large batch training with stochastic gradient descent (SGD).
\subsection{RNN-T model architecture}
\label{sec:rnnt}
\begin{figure}[htp]
	\centering
	\includegraphics[width=0.48\linewidth]{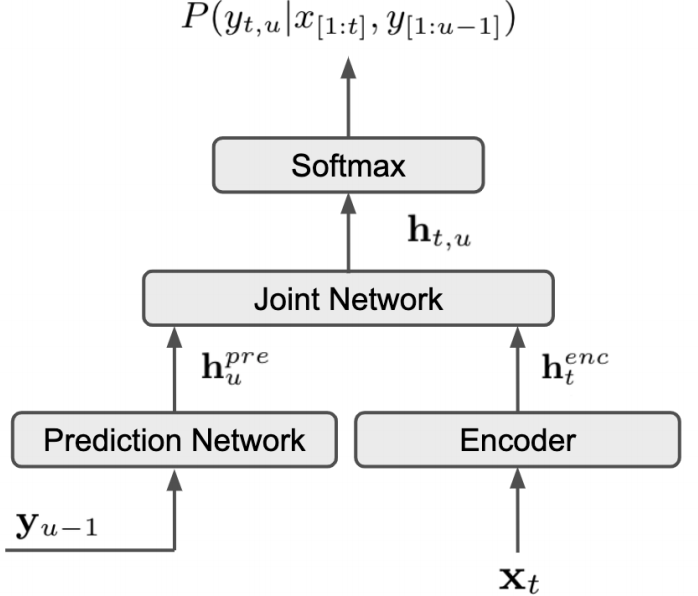}
	\caption{RNN-T ASR model architecture}
	\label{fig:rnnt} 
\end{figure}

Figure \ref{fig:rnnt} shows the RNN-T~\cite{graves2012sequence} architecture used in real-time speech recognition. The model predicts the probability $P(\textbf{y}|\textbf{x})$ of labels $\textbf{y}=(y_{1},...,y_{U})$ given acoustic features $\textbf{x}=(x_{1},...,x_{T})$. It has an encoder, a prediction network, and a joint network. The encoder is analogous to an acoustic model that takes a sequence of acoustic input features and outputs encoded hidden representations. The prediction network corresponds to a language model that accepts the previous output label predictions, and maps them to hidden representations. The joint network is a feed forward DNN that takes both the encoder and prediction network outputs, and predicts the final output label probabilities with softmax normalization.
\subsection{Overview of learning with large batch size}
\label{sec:large-batch-overview}
\noindent When training with SGD, mini batches with a well crafted decaying learning rate schedule are commonly used as opposed to using large batches. Previous work in \cite{keskar2016large} has demonstrated a generalization drop when using large batches, thus recommending mini-batch SGD with decaying learning rate. However, recent advances in large batch training both with a linear scaling rule of the learning rate \cite{goyal2017accurate} and constant learning rate\cite{smith2018dont}, large batch training has been shown to achieve similar performance as its mini-batch counterpart. A recurrent observation in the literature~\cite{keskar2016large,LI:KDD,shallue2018measuring,mccandlish2018empirical,golmant2018computational,masters2018revisiting} is that large batch training (for ImageNet, $> 1000$) results in test accuracy degradation. Despite the warm-up in \cite{goyal2017accurate}, for ImageNet, the best accuracies are observed up to a large mini-batch of $8192$ images.  

In this paper, we deal with the challenges of 1) training with large batches in incremental learning and 2) semi-supervised learning to alleviate unavailability of human annotation and labels. For automatic speech recognition (ASR), with large batch sizes ($> 3e5$ utterances) using a fixed learning rate schedule, we observe better test accuracies, as opposed to the degradation in literature, while training with teacher transcripts for the incremental audio data.
\section{ILASR: Incremental Learning for Automatic Speech Recognition}
\label{sec:flc-training}
This section describes the ILASR architecture and the corresponding incremental learning algorithm. \textsc{ILASR} offers large scale end-to-end ASR training with the ability to incrementally update the models in user-defined time windows. ILASR automates the whole life-cycle of data generation, sampling, labeling, model development, evaluation and deployment for audio data in near real-time. 
\begin{figure}[htp]
	\centering
	\includegraphics[width=0.9\linewidth]{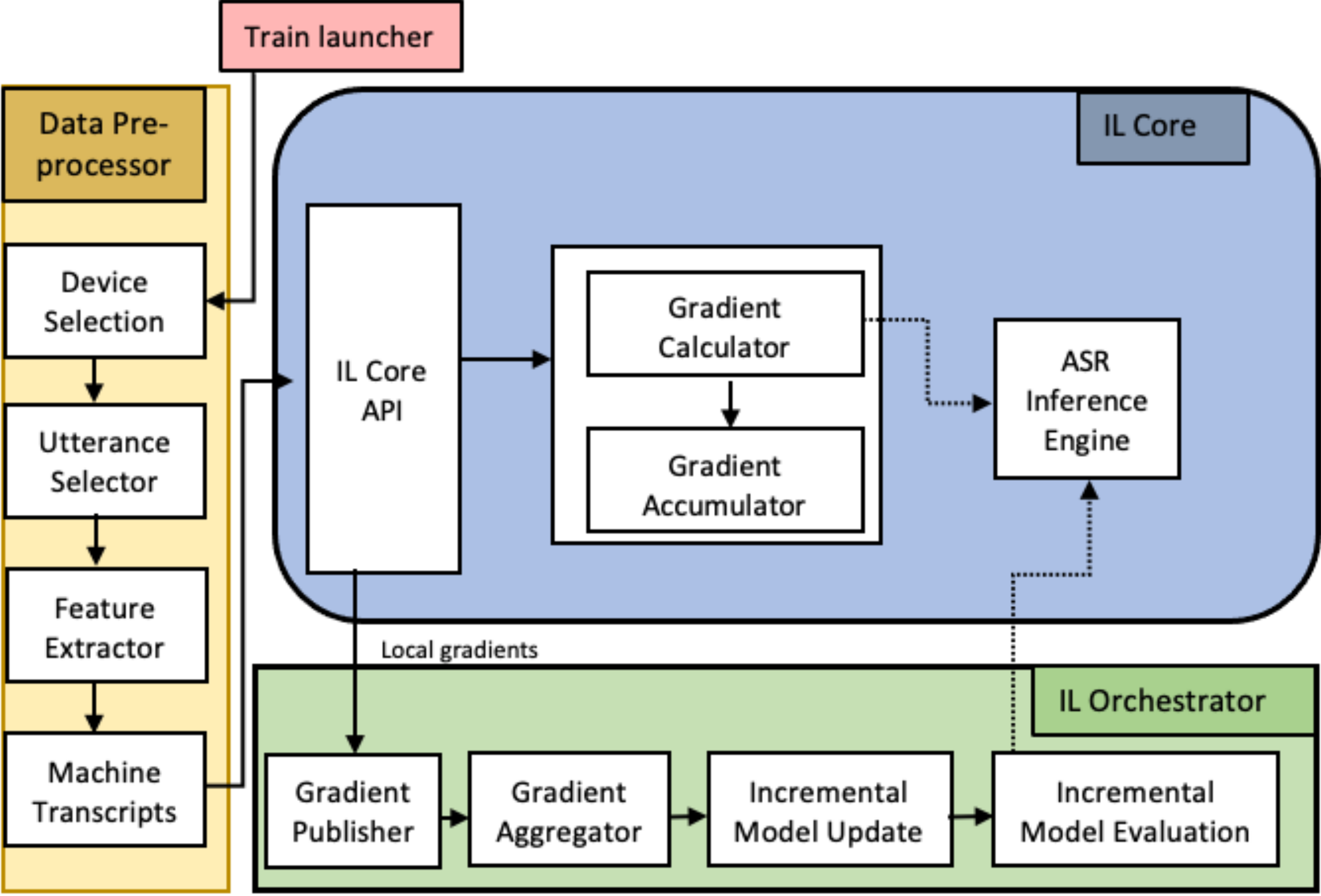}
	\caption{High-level skeleton of the ILASR architecture}
	\label{fig:flc_arch} 
\end{figure}

\subsection{ILASR Architecture}
\label{flc-arch}
Figure~\ref{fig:flc_arch} shows the architectural overview of ILASR system. The system comprises three primary components: (1) {\em Data preprocessor} -- is a cloud runtime service that processes near real-time audio from device; (2) {\em IL Core} is responsible for model training, computing model updates and inference; and (3) {\em IL Orchestrator} aggregates the accumulated gradients, updates the model, performs evaluation and finalizes the model update based on the evaluation result.

\noindent {\em Train launcher} initiates the end-to-end ASR training in ILASR. The first step is {\em data preprocessing} to select a subset of devices and utterances to participate in the training loop. The selection could be random or based on heuristics aimed at improving the model in a particular way. Confidence scores obtained during inference are used ~\cite{jiang2005confidence,kalgaonkar2015estimating} coupled with heuristics such as presence of rare words or semantic tags and intents of interest. This selection can be extended to leverage weak signals from user feedback such as user indicating whether the action taken by the assistant is positive or negative or detecting friction such as repeated requests or cancellations. Acoustic features are extracted and augmented~\cite{park2019specaugment} for the selected utterances for training. {\em Machine transcripts} are generated using a teacher ASR model pre-trained using standard distributed training. The Conformer~\cite{gulati2020conformer} based end-to-end ASR teacher model decodes the input audio ($X$) to produce machine transcripts ($Y$). These paired ($X$, $Y$) instances are used to train the model.
The machine transcripts act as ground truth labels. ILASR produce transcriptions through secure automation without human intervention or review. The extracted features together with machine transcripts in this step are combined to train the student models using {\em IL Core}.\\
\noindent The {\em IL Core} system has an application programming interface (API) that supports local gradient accumulation on each of the servers in the fleet, and an ASR inference engine. The IL Core API supports FedSGD and FedAvg~\cite{mcmahan2017communication} and can be extended to support other federated optimizers such as FedProx~\cite{sahu2018convergence}, FedMA~\cite{wang2020federated}, FedNova~\cite{wang2020tackling}, and adaptive federated optimizer~\cite{reddi2020adaptive}. 
\noindent The {\em IL Orchestrator} coordinates training across the ILASR fleet. IL Orchestrator contains the gradient publisher, aggregator and updates the model incrementally. The gradient aggregator collects gradients from each of the IL Core instances, aggregates them and then applies them to the current model. Once the model update is done, the collected gradients are discarded and not stored in the system which helps with reducing the risk of gradient inversion attack. A periodic light-weight evaluation of the model ensures that the model is directionally improving. The global model is updated in a given round when the performance improves over that of the previous round. To reduce the probability of a model update resulting in worse performance, ILASR can be run in parallel with differing hyperparameters. In this scenario, one of the resulting models can be utilized should it result in improved performance. After a sufficient number of rounds, the final model is stored for the next model release after a detailed model validation step.

\noindent ILASR addresses security and privacy concerns with different levels of granularity. Since ILASR is a cloud-based system for privacy-preserving IL at scale, the audio encryption is two fold. In the first stage, TLS~\cite{dierks2008transport} encryption is applied on audio transmission followed by an application level key-master~\cite{mahajan2013study} encryption. Importantly, the audio is purged in a few minutes ($\le 10$), within which the model updates are calculated.

\subsection{ILASR: Incremental Learning}
\label{sec:ILASR-il}
\begin{algorithm}\small
	\caption{ILASR incremental learning algorithm}\label{alg:ILASR-il-alg}
	\begin{algorithmic}[1]
		\REQUIRE $\mathcal{K}$ servers, $\mathcal{L}$ loss function, $N$ number of local steps per round, $\mathcal{B}$ local batch size, $(\eta)$ learning rate, $P_{k}^{r}$ recent utterances pulled by server $k$ in round $r$, $\mathcal{D}_{eval}$ eval set and $\mathcal{D}_{ht}$ past transcribed data if used for rehearsal training.
		\ENSURE $w_\mathcal{G}^{r}$ incrementally updated global model and $wer_{r}$ word error rate on the eval set after $r$ rounds
		\STATE Init. $w_\mathcal{G}^{0}$// start training with a pre-trained model
		\STATE $wer_{0} = asr\_inference\_engine(\mathcal{D}_{eval}, w_\mathcal{G}^\mathcal{T})$
        \FOR{each round $r = 1, 2, \dots$}
        \FOR{each server $k \in$ ILASR Fleet \textbf{in parallel}}
        \STATE $w_{k}^{r} = w_\mathcal{G}^{r-1}$
        \STATE  $\mathcal{D}_{ssl} \leftarrow$ (filter $P_{k}^{r}$ based on utterance selection criteria  and generate machine transcript, refer algorithm \ref{alg:cutoff})
        \STATE $\mathcal{D}_{train} \leftarrow$ (mix $\mathcal{D}_{ssl}$ and $\mathcal{D}_{ht}$ if $\mathcal{D}_{ht}$ is used for rehearsal, else just $\mathcal{D}_{ssl}$)
        \STATE  $\mathcal{D}_{train} \leftarrow$ ( split $\mathcal{D}_{train}$ into $N$ batches of size $\mathcal{B}$)
        \FOR{each batch $b_{i}$ from $b_{1}$ to $b_{N}$}
        \STATE $w_{k}^{r} \leftarrow$ $\textrm{optimizer}_{k}$.update($\eta, \nabla \mathcal{L}(w; b_{i})$)
        \ENDFOR
        \ENDFOR
        \STATE $w_\mathcal{G}^{r} \leftarrow \frac{1}{\mathcal{K}}\sum_{k=1}^{\mathcal{K}} w_{k}^{r}$
		\STATE $wer_r \leftarrow asr\_inference\_engine(\mathcal{D}_{eval}, w_\mathcal{G}^{r})$
		\STATE $w_\mathcal{G}^{r} = w_\mathcal{G}^{r-1}$ if $wer_r > wer_{r-1}$ // Revert to the previous model if not a better model.
		\ENDFOR
	\end{algorithmic}
\end{algorithm}

\noindent Algorithm~\ref{alg:ILASR-il-alg} shows the incremental learning policy in ILASR framework. The new model obtained in each round is used only if it performs better than the model from the previous round. Parallel runs of the algorithm with differing hyper parameters to train an ensemble of incrementally updated models can ensure that there is at least one model that performs better than the model from the previous round. Another interesting consideration is the effect of catastrophic forgetfulness~\cite{mccloskey1989catastrophic,french1999catastrophic,goodfellow2013empirical} in incremental learning of ILASR framework, where the previous learned behaviour of a model is forgotten with new updates. This can be mitigated with the rehearsal~\cite{robins:forgetting} of training on a subset of annotated historical data along with the new data.

We describe the SSL data generation method in algorithm~\ref{alg:cutoff}. We randomly sample a subset of the audio in near real-time, to prepare a data pool ($\mathcal{P}$), and calculate target number of utterances ($\mathcal{U}$) to be sampled from $\mathcal{P}$, where each of the utterances include a pre-calculated confidence value \cite{swarup2019improving}. For each confidence bin, for example confidence in (600,700] where confidence is evaluated on a scale from 0 to 1000, utterances are filtered to conform to the confidence criterion.
The randomly sampled utterances from above are set to get target number of utterances and sent to IL core for training, which are deleted as soon as the model takes a pass over it for the first time. Additional criteria such as presence of rare words, presence of desired semantic tags can also be utilized. 
\begin{algorithm}
	\caption{SSL data selection procedure}\label{alg:cutoff}
	\begin{algorithmic} [1]
		\REQUIRE $\tau$ list of utterance confidence bins 
		\ENSURE $\mathcal{X}$ data set
		\STATE $\mathcal{P}=random\_sample()$ // prepare a random pool of data
		\STATE $\mathcal{P}=teacher\_decode(\mathcal{P})$  // generate machine transcripts
		\STATE $\mathcal{U}=calc\_utterance\_confidence(\mathcal{P})$
.		\STATE $\mathcal{Q} = [\; ]$ // a bin for each confidence range
		\FOR{$c \in \tau$}
		\STATE $\mathcal{Q}[c] = filter\_utterances(\mathcal{U}, c)$
		\STATE $\mathcal{X} = select\_utterances(\mathcal{Q})$ // can include additional criteria like presence of rare words or desired semantic tags
		\ENDFOR
	\end{algorithmic}
\end{algorithm}

\section{Experiments}
\label{sec:experiments}
We describe the datasets, model configurations and experimental settings used in this paper, to provide insights and study privacy-preserving incremental learning through ILASR.

\subsection{Datasets}
\label{sec:datasets}
All speech data used for training and evaluation are de-identified.

\noindent \textbf{Train sets}
The audio streams are prepared into offline training datasets. The following training datasets are used for experimentation:

\noindent\textbf{Pre-training datasets}: A $480k$-hour pre-training dataset is utilized for building pre-training models. This pre-trained model is used as a starting point for incremental training with the ILASR system. This comprises two datasets:
\begin{enumerate}
   \item \textit{120K-hour HT}: Human-transcribed (HT) data from $2020$ and previous years
    \item \textit{360K-hour SSL}: Machine-transcribed data in $2020$
\end{enumerate} 

\noindent\textbf{Incremental training dataset}: We consider the end of $2020$ as the start date for incremental training of ASR models.

\begin{enumerate}
    \item \textit{180K-hour ILASR SSL}: Machine-transcribed data is generated over a period of six months in $2021$ (Jan to June) and is used for near real-time training of the ILASR system. 
\end{enumerate} 

\noindent \textbf{\text{Test sets}}: We evaluate the models on in-house human transcribed (HT) test sets.\\
\noindent\textbf{\textit{General}}: Includes three HT datasets from different time ranges representing the general use case. It  comprises a $37$-hour test set from $2021$, a $10$-hour test set from $2020$ and a $96$-hour test set from $2018-2019$.\\
\noindent\textbf{\textit{Rare}}: Includes three HT datasets from different time ranges, where the transcriptions contain at least one rare word. Rare words are those in the long-tail of the vocabulary determined by word frequency. This includes a $44$-hour test set from $2021$, a $44$-hour test set from $2020$, and a $27$-hour test set from $2018-2019$.\\
\noindent\textbf{\textit{Delta}}: This consists of a $22$-hour HT test set that records a change in frequency of words in $2021$ over $2020$. The transcriptions are filtered based on 1-gram, 2-gram and 3-grams that are 5x more frequent in $2021$ than $2020$. This test set captures changes in the data distribution and is very relevant to measure the impact of incremental learning with ILASR.\\
\noindent\textbf{\textit{Messaging}}: Includes two HT datasets that comprise of messaging and communications domain data. It includes a $2.7$-hour HT test from $2020$ and a $45.5$-hour HT test set from $2018-2019$.\\
\noindent\textbf{\textit{Monthly datasets (2021)}}: We use six monthly test sets from Jan to June $2021$ to evaluate the incremental learning setup of ILASR. Each of these datasets are refered to as (Jan, Feb,$\cdots$,June) and each month has on average $70$-hours of data. We further report results on 3-month datasets $Jan-Mar$ including data from Jan, Feb, Mar and $Apr-Jun$ including data from Apr, May, June.

\subsection{Model details}
\emph{\textbf{Features}}: The audio features are $64$ dimensional log-mel filter-bank energies ~\cite{nadeu1995decorrelation} computed over a $25$ms window, with a $10$ms shift. The features computed on $3$ consecutive $10$ms frames are stacked and sub-sampled to result in $192$ dimensional features at a $30$ms frame rate, and are provided as input to the ASR model. The ground truth transcripts are tokenized to $2500$ sub-word units using a uni-gram language model~\cite{Kudo2018SubwordRI}.

\noindent \emph{\textbf{Models}}:
\label{sec:models}
\textit{Teacher models}: 
Teacher models are used to generate SSL machine transcripts. We have three teacher models available: $T3$ is a teacher model (a conventional RNN-HMM hybrid ASR system\cite{bourlard2012connectionist}) that is trained on $100K$-hours of data until $2019$ only. The machine-transcripts from $T3$ are utilized to bootstrap and provide transcripts for the more recent $360K$-hour SSL pre-training dataset. The $480K$-hour pre-training dataset, including the $360K$-hour SSL dataset based on $T3$ and the $120K$-hour HT dataset, is utilized to train two updated teacher models: (1) $T1$: A larger conformer based ASR architecture~\cite{gulati2020conformer} trained on $480K$-hours. $T1$ has ~122M parameters, an encoder with $17\times512$ LSTM layers, $8$ attention heads with $32$ dimensional convolution kernel. The prediction network uses $2\times1024$ LSTM layers. (2) $T2$ is a conventional RNN-HMM hybrid ASR system \cite{bourlard2012connectionist} and is trained on the same $480K$-hour dataset.
Finally, the student models for all experiments in the paper are trained on SSL datasets that use the most recent $T1$ teacher model. In section \ref{sec:teacher-ablations}, for the purpose of ablations comparing various teachers, we train student models on SSL datasets that are based on $T2$ and $T3$.

\noindent\textit{Student models}: 
The student models are based on different LSTM based RNN-T architectures. These vary in the number of encoder layers and the feature frame rates. Two student models are described as follows. {\em \underline{$rnnt\_60m$}} contains ~60M parameters with $5\times1024$ LSTM encoder, $2\times1024$ LSTM prediction network and a feed-forward joint network with {\em tanh} activation. The input embeddings of the prediction network are $512$ dimensional. SpecAugment~\cite{park2019specaugment} is used on the audio features. {\em \underline{$rnnt\_90m$}} contains ~$90$M parameters with $8\times1024$ LSTM layer encoder, a prediction network of size $2\times1024$, and a feed-forward joint network with {\em tanh} activation. The input embeddings of the prediction network use $512$ dimensional embeddings and a $2500$ sub-word tokenizer from a uni-gram language model. SpecAugment is used on the audio features. The encoder uses an LSTM based time-reduced~\cite{Soltau2017ReducingTC} RNN multi-layer (for speed of training and inference) with feature frame rate set to $3$ layers. Each of these feature frame layers have $1536$ units and the LSTM projection with a size of $512$.

The models $rnnt\_90m$ and $rnnt\_60m$ are pre-trained on both the HT data of $120K$ hours and $340K$ hours of SSL data generated using the teacher ($T1$) decoded labels. The human transcribed data used in the pre-training utilizes data up to the end of $2020$, while the SSL data is in $2020$. For our experiments in this paper, we further train the above pre-trained RNN-T student models using a total amount of $180k$ hours of SSL data (teacher generated labels) available in a time-window of $6$ months in $2021$.

\noindent\textbf{Training details}: We use the following parameters to train both the teacher and student models. The system is run on a fleet consisting of $200$ nodes.  We adopt a learning rate schedule of warm-up where $lr=1e-7$ for the first $3000$ steps, followed by constant learning rate of $5e-4$ till $50k$ steps, then exponential decay ($lr=1e-5$) from $50k$ to $750k$ steps with Adam optimizer (hyperparameters are $\beta_1=0.9$, $\beta_2=0.99$). 

We experiment with multiple large batch sizes ($9k$, $18k$, $73k$, $147k$, $215k$, $307k$) through gradient accumulations. Note that these accumulations have an implicit effect of changing the gradient values due to the summation of gradients across a large batch. We process large batches without altering the $lr$ schedule while accumulating the gradients.
The performance of these models is measured in terms of relative word error rate reduction (WERR) over the corresponding baselines. WER is the ratio of edit distance to sequence length, where edit distance is the length of the shortest sequence of insert, delete and substitution operation on transforming a predicted sequence to target. 
\section{Results \& Discussion}
\label{sec:results-discussion}
In this section, we analyze the performance of incremental learning in ILASR. In particular, we analyze the performance of incremental learning in ILASR in terms of relative word error rate reduction (WERR) in comparison with the initial pre-trained student models as baselines. 

\begin{table}[htp]\small
	\centering
	\caption{Relative \% WER improvements from the initial model when trained with the ILASR system}\label{tab:replay-vs-no}\small
	\begin{tabular}{|c|c|c|c|}
		\toprule
		& & \multicolumn{2}{c|}{\textbf{ILASR}}\\
		\cline{3-4}
		\textbf{Time} & \textbf{Test-set} & \textbf{replay} &\textbf{no replay} \\
		\midrule
		\multirow{5}*{2021} & Rare & 0.72\%   & 0.66\%   \\
		 & Delta     & 20.10\%   & 23.99\%      \\ 
		 & General & 1.23\%  & 0.41\%  \\
		& Jan-Mar    & 1.25\%    & 1.50\%       \\ 
		& Apr-Jun & 2.73\%  & 3.09\%  \\
		\midrule
		\multirow{3}*{2020} & Rare & 0.62\%  & 0.62\%  \\
		& General & 0.00\%  & -0.72\%  \\
		& Message & -0.83\%  & -2.04\%  \\
		\midrule
		\multirow{3}*{2018-2019} & Rare & -0.63\%  & -0.63\%  \\
		& General & -1.21\%  & -2.6\%  \\
		& Message & -2.82\%  & -3.42\%  \\
		\bottomrule
	\end{tabular}
\end{table}

From Table \ref{tab:replay-vs-no}, we see that ILASR improves a strongly trained base model by up to 3\% on test sets in 2021 which climbs to 20\% on the delta dataset that consists of new or trending words and phrases. At the same time, performance on older general and tail test sets do not see much degradation. 

Catastrophic forgetting is one of the issues incremental learning needs to circumvent in order to have consistent performance across both old and new data. In Table \ref{tab:replay-vs-no}, we compare the performance of \emph{replay} based incremental learning, where a sub-sampled portion of 120K-hour human-transcribed data is also consumed in model training while the \emph{no replay} counterpart does not involve that. As demonstrated in Table \ref{tab:replay-vs-no}, \emph{replay} based training tends to outperform its \emph{no replay} counterpart on older test sets as expected from IL literature.

\begin{figure}%
    \centering 
    \includegraphics[width=0.9\linewidth]{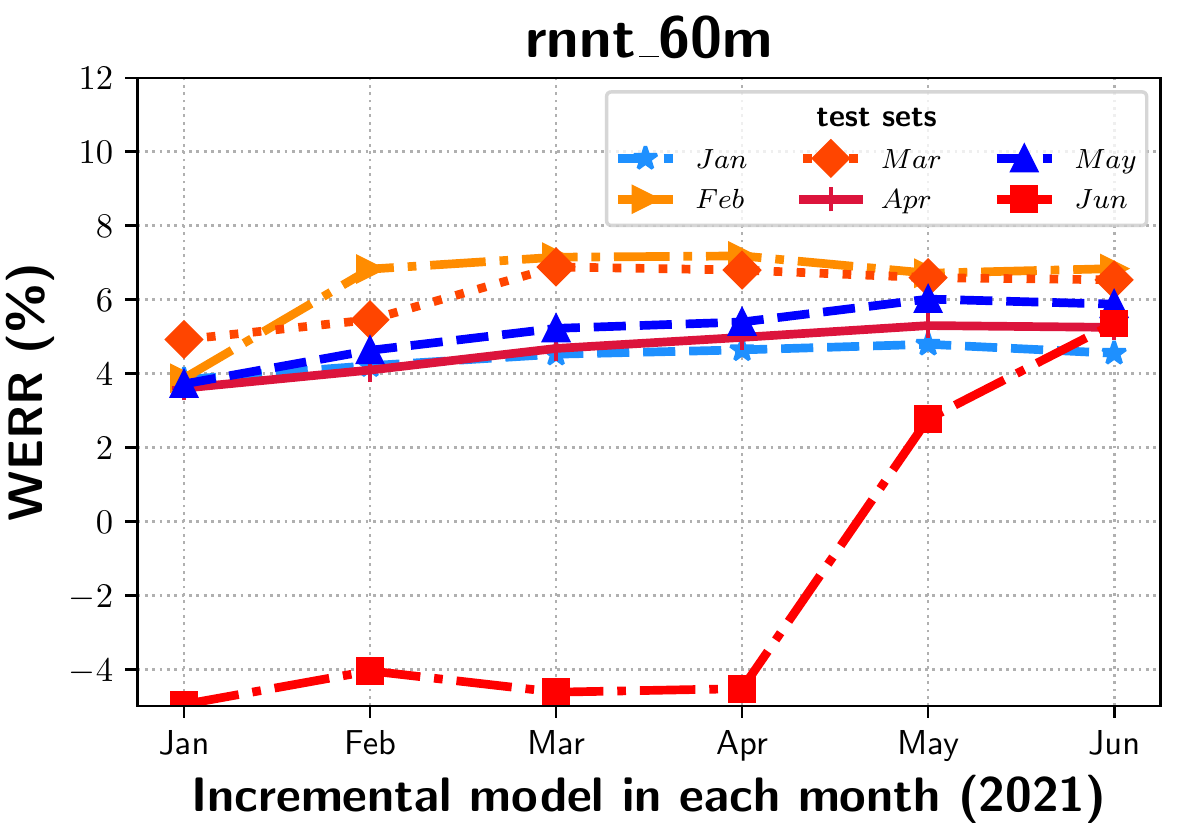}
    \caption{Monthly WERR (\%) for incremental learning in ILASR for $rnnt\_60m$ on six of the monthly test sets (${Jan}$--${Jun}$) when measured relative to the starting model versus the one trained incrementally in each month.}\label{fig:il-flc-incremental}%
\end{figure}
\begin{figure*}[htp]
	\centering
	\begin{subfigure}{.31\textwidth}
		\centering
		\includegraphics[width=0.9\linewidth]{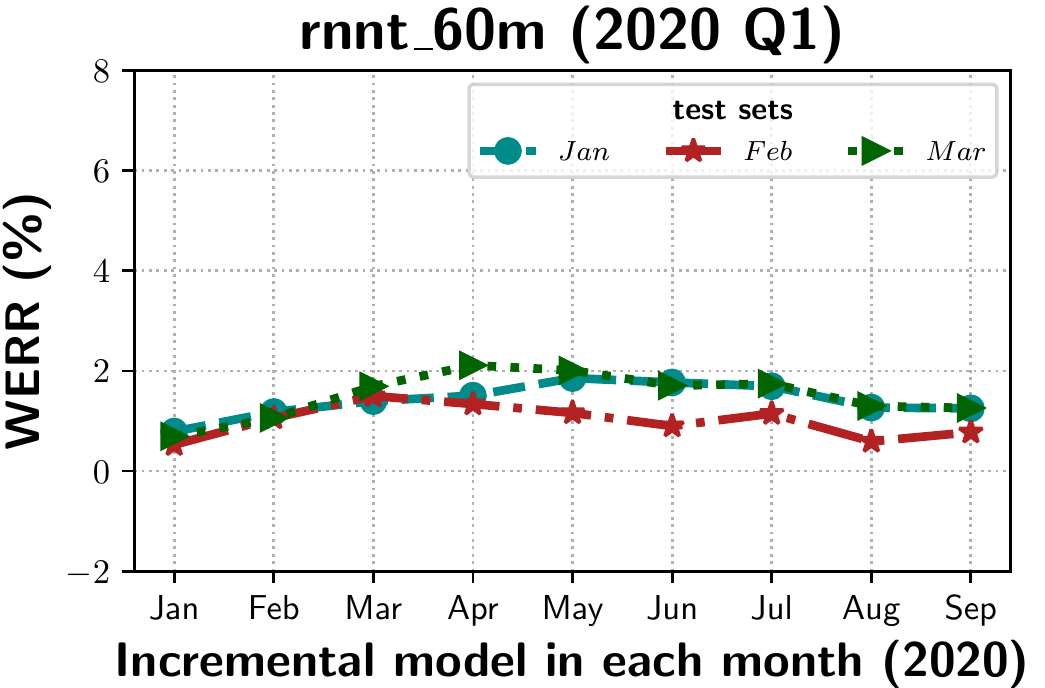}
		\caption{$2020 Q1$}\label{fig:old-inc-q1}
	\end{subfigure} \hfill
	\begin{subfigure}{.31\textwidth}
		\centering
		\includegraphics[width=0.9\linewidth]{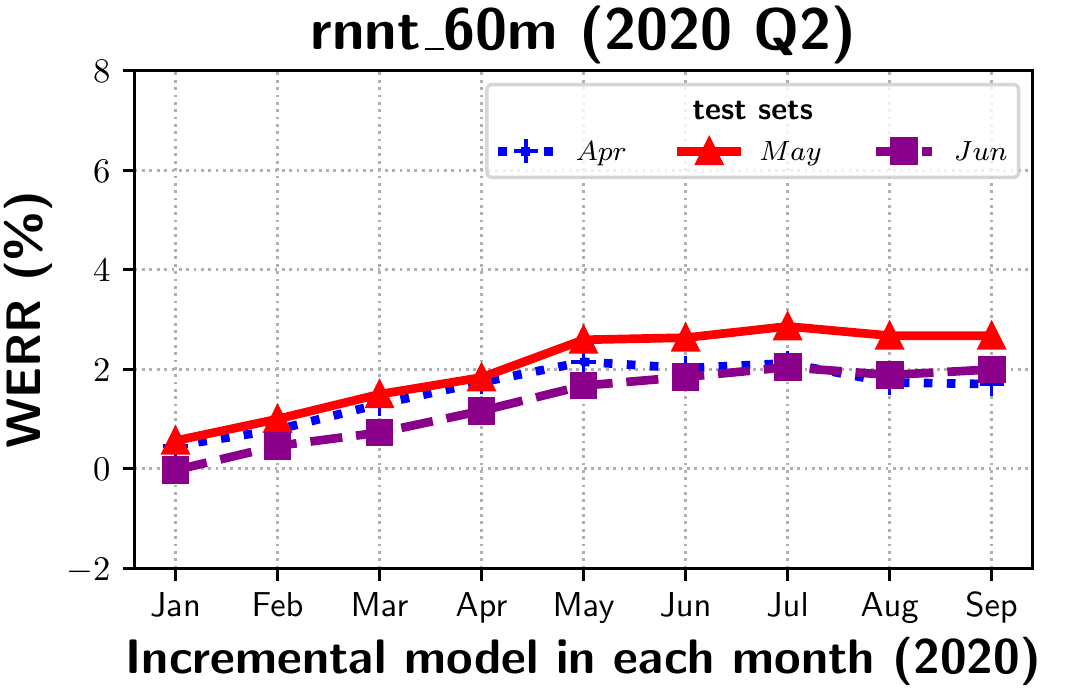}
		\caption{$2020 Q2$}\label{fig:old-inc-q2}
	\end{subfigure}\hfill
	\begin{subfigure}{.31\textwidth}
		\centering
		\includegraphics[width=0.9\linewidth]{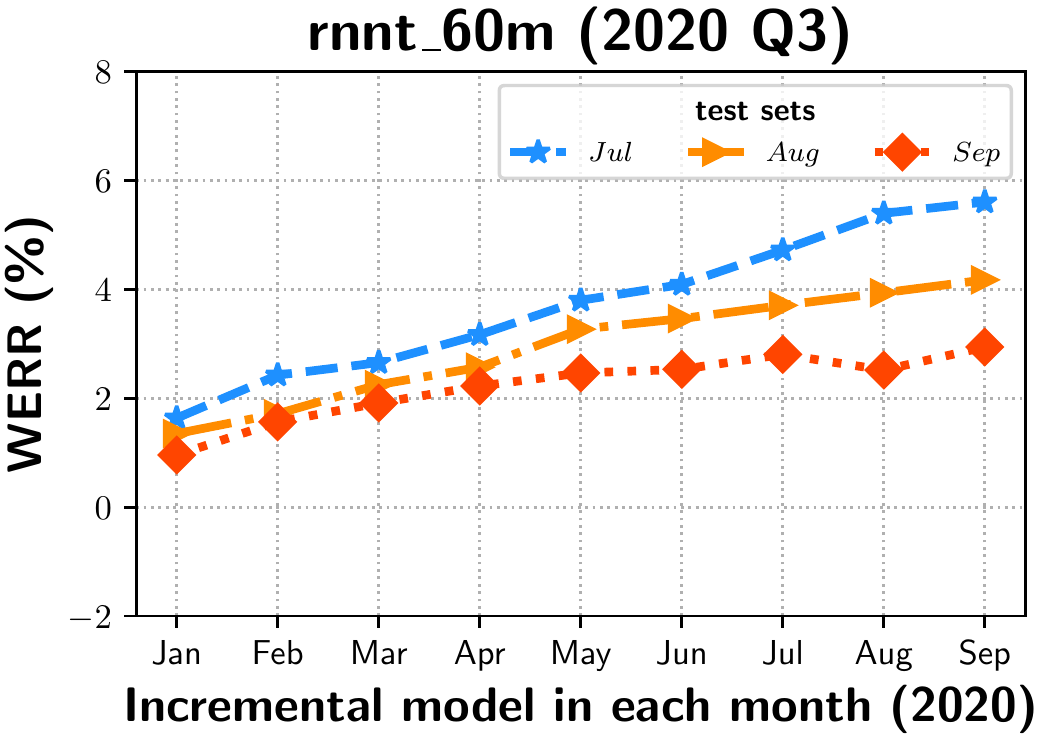}
		\caption{$2020 Q3$}\label{fig:old-inc-q3}
	\end{subfigure} 
	\caption{For the $rnnt\_60m$, the pre-trained model is trained on the data available until $12/2019$. Training this pre-trained model in incremental mode for the next nine months ($Jan-Sep$) in $2020$. The x-axis shows the monthly incremental model, where the model from previous month is fine-tuned on the data in current month; y-axis shows the relative WER in each month w.r.t the initial pre-trained model. Each of the curves represent the test set of the corresponding month.}\label{fig:old-inc}
\end{figure*}

Next, we evaluate the incrementally trained ILASR models on fine-grained test sets that are prepared in each of the six months (Jan-Jun) of $2021$, see Figure~\ref{fig:il-flc-incremental}. For all the evaluations in Figure~\ref{fig:il-flc-incremental}, we report the WERR in each month relative to the initial pre-trained model (for example, WERR in $May$ is the relative difference between the WERs of $May$ model and the pre-trained). The results show incremental improvements in performance on all the six monthly test sets from month to month in the ILASR training. This suggests that the incremental training helps in capturing the new trends in time periods while the model is adapting to the incremental changes in the data. It is also noteworthy that the incremental improvement does not come at the cost of catastrophic forgetting. More interestingly, the models trained with the data until $May$/$June$ degrade the performance on $June$ test-set, which improves after the model is trained on the data available from $May/June$. This clearly suggests the adaptive nature of capturing the shifts in data in the new time periods in ILASR. 

To further strengthen the incremental learning claims, we analyze the incremental learning patterns for a longer duration in the time-periods between $Jan-Sep$ in $2020$. Figure~\ref{fig:old-inc} shows the learning patterns on a quarterly basis for the first three quarters (Q1--Q3) of 2020. In $2020Q1$ and $Q2$, the WERR improves initially and then decreases as the incremental model training progresses on a month-over-month basis. The degradation (whilst better than the baseline) is a demonstration of forgetting as newer updates are prioritized over the months old test sets. Consequently, in $2020Q3$, the performance improves without any downward trends, which is due to the fact that the models keep learning month over month while the test sets also belong to the same time-periods. These trends suggest that the proposed techniques help in incrementally improving the performance even in longer time-periods while limiting the regressions on the older eval data.       

Next, we explore several design choices which play a key role in the performance of ILASR and share our insights in terms of the design choices.

\subsection{Design Choices: ILASR}
\label{design-choice}
We explore the following design choices in the context of the ILASR framework:  1) effect of large batch sizes on performance of the student models; 2) temporal effects on processing the data in ILASR; 3) analyze the importance of  different teacher models in ILASR.
\begin{table}[htp]\footnotesize
	\centering
	\caption{Effect of large batches on the relative improvement in performance (in terms of WERR, $\%$) of all the three models when fine-tuned in ILASR. }\label{tab:large-batch}
	\begin{tabular}{|c|c|c|c|c|c|c|c|}
		\toprule
		\multirow{3}{*}{\textbf{Time}} & \multirow{3}{*}{\textbf{Test-set}}        & \multicolumn{6}{|c|}{\textbf{effective batch size}}   \\ 
		& & \textbf{9k}   & \textbf{18k} & \textbf{73k}   & \textbf{147k}        & \textbf{215k}  & \textbf{307k}     \\ 
		\cline{3-8} 
		\cline{3-8} 
		& &  \multicolumn{6}{|c|}{\textbf{$rnnt\_60m$}}   \\ 
		\midrule
		2021 & Jan--Mar  & 2.74\%     & 1.49\%     & 2.37\%     & 2.37\%     & 2.24\%     & 2.24\%     \\ 
		\midrule
		2018--& Rare    & 3.58\%      & 3.58\%   & 3.72\%    & 3.65\%    & 3.78\%    & 3.72\%    \\ 
		2019 & Message    & 6.46\%    & 6.05\%    & 9.46\%    & 9.46\%    & 9.28\%    & 9.37\%    \\ 
		& General & 15.14\%   & 15.12\%   & 14.70\%    & 14.56\%     & 14.41\%     & 14.26\%     \\ 
		\bottomrule
	\end{tabular}
\end{table}

\subsubsection{Training is robust to large batch sizes}
We use large batches in ILASR via gradient accumulations. As the effective batch size increases, the number of optimization or update steps reduces as the same amount of data is processed. Larger batch sizes would require fewer optimization steps and vice versa for the same amount of data. Use of large batches accelerates the training (shown in~\cite{you2017scaling}), which is similar in ILASR. The reason large batch sizes is relevant in the ILASR system is that there are limitations about how quickly gradients can be aggregated and the global model distributed to the servers in the fleet. Hence, a limited number of update steps can take place in a time period compared to GPU-based offline distributed training. Moreover, as data arrives in a streaming fashion and is not persisted, it needs to be consumed as and when it arrives, in near real-time. For each of the limited number of updates, a large amount of streaming data is available.  \\
\noindent We explore the trade-off between large batches and model performance. Table~\ref{tab:large-batch} shows the effect of large batches on performance of a student models trained in ILASR. The performance (WERR) is relative to the corresponding pre-trained student model. This baseline is weaker, hence improvements are larger. We find that increasing the batch multiplier (effective batch size) has insignificant effect on WER. As batch sizes increase from 9K to 300K utterances, the difference in the accuracies is insignificant. \\
\noindent More importantly, this finding is in contrast to the test accuracy degradation effects reported in literature~\cite{keskar2016large,LI:KDD,goyal2017accurate,shallue2018measuring,mccandlish2018empirical,golmant2018computational,masters2018revisiting,sun2019optimizing} with the use of large batches. We observe that such degradation is not evident for model training in ILASR. Although, the attempts in the literature have no strong mathematical justification, Goyal et al.~\cite{goyal2017accurate} reasoned the performance degradation to optimization issues, thereby using warm-up to mitigate the degradation. 
\begin{table}[htp]\small
	\centering
	\caption{Impact of the temporal order (chronological versus random) of processing the training data in ILASR for both with and without replay of the human transcriptions.}\label{tab:zero-day}
	\begin{tabular}{|c|c|c|c|}
		\toprule
		\textbf{Time} & \textbf{Test-set} & \multicolumn{2}{c|}{\textbf{Chrono vs. random}}\\
		\cline{3-4}
		& & \textbf{replay} & \textbf{no replay} \\ 
		\midrule
		\multirow{4}*{2021} & Rare & -0.62\%  & -1.16\%  \\
		& Delta & -1.68\%  & -0.73\%  \\
		& General & 0.15\%  & 1.47\% \\
		& Jan-Mar  & -0.53\%  & -0.24\%   \\
		& Apr-Jun & -0.47\%  & 0.29\%  \\
		\midrule
		\multirow{3}*{2020} & Rare & -0.56\%  & -0.90\%  \\
		& General & -0.55\%  & 1.61\%  \\
		& Message & 0.35\%  & 0.48\%  \\
		\midrule
		\multirow{3}*{2019-2019} & Rare & -0.46\%  & -1.26\%  \\
		& General & 0.32\%  & 0.67\%  \\
		& Message & -0.11\%  & -0.87\%  \\
		\bottomrule  
	\end{tabular}
\end{table}
\noindent Similarly, in our case, we attribute the gains and/or no performance degradation to the following factor. The initialized models are pre-trained that have converged on the data from a previous time period as opposed to random initialization in the large batch training in literature, usually, these models are trained from scratch (despite the few initial epochs in warm-up) in the literature.
\subsubsection{Impact of chronologically ordered data}
One important aspect of IL is the data being processed in time as is available, chronologically. We analyze the effect of processing order (chronological vs random) for the six months in $2021$. Note, random order is same as shuffling the data in regular distributed training of deep models. Chronological data is not IID across time as utterances have a correlation with the time of day (for example, requests to snooze alarms in the morning or turning smart lights on after sundown). We found that there is no difference in performance of processing the data chronologically as compared to randomly as depicted in Table~\ref{tab:zero-day}. Moreover, in both the cases of chronological and randomized, the improvements over initial baselines are clearly evident (see Table~\ref{tab:replay-vs-no}).
\subsubsection{Ablations with teachers and students}
\label{sec:teacher-ablations}

\begin{table}[htp]
	\centering
	\caption{Performance (in terms of WERR, $\%$) of the RNN-HMM hybrid ASR teacher ($T2$) and bidirectional RNN-HMM hybrid ASR ($T3$) based teacher models with respect to the Conformer teacher ($T1$). The negative (-) sign represents that $T1$ performs worse while the rest shows that $T1$ is the best performing teacher model.}\label{tab:teachers-comp}
	\begin{tabular}{|c|c|c|c|}
		\toprule
		Time & Test-set   & $T1$ vs $T3$  & $T1$ vs $T2$  \\
		\midrule
	    2021 & Jan--Mar & $16.63\%$  & $0.14\%$  \\
	    \midrule
		\multirow{3}*{2018--2019} & Rare & $8.75\%$   & $12.02\%$ \\
		& Message & $7.34\%$   & $14.92\%$ \\
		& General & $-0.89\%$    & $20.51\%$\\
		\bottomrule
	\end{tabular}
\end{table}

We experiment with three different teacher models that are trained for different time ranges with different architectures. This experiment helps us explore the importance of keeping an updated and more effective teacher. The three teachers are: $T1$ is the Conformer based that is explained earlier in section~\ref{sec:models}; $T2$ is a RNN-HMM conventional hybrid model \cite{bourlard2012connectionist}; $T3$ is a bidirectional RNN-HMM conventional hybrid ASR model. $T1$ and $T2$ are trained on the same amount of data until the end of $2020$ while $T3$ is trained on the data (a total of $\sim100k$ hours of HT data) available till the end of $2019$. 

Table~\ref{tab:teachers-comp} compares the performance of the teacher models. On an average, $T1$ is better than the rest of the two teachers, $T1>T2>T3$ on new data reflecting the importance of keeping the teacher model up-to-date. Conformer based teacher, $T1$ is better than the rest of the remaining two teachers. The relative performance differences, when measured on the four standard test sets are, $T1$ is better than $T2$ and $T3$ with $11.85\%$ and $7.96\%$ WERR, respectively.

\begin{table}[htp]\footnotesize
	\centering
	\caption{The performance (in terms of WERR) of the student models when trained with the machine transcripts generated from each of the three different teacher models.} \label{tab:teachers}
	\begin{tabular}{|c|c|c|c|c|c|c|c|}
		\toprule
		\multirow{2}{*}{\textbf{Time}} & \multirow{2}{*}{\textbf{Test-set}} & \multicolumn{3}{|c|}{$rnnt\_90m$}  & \multicolumn{3}{|c|}{$rnnt\_60m$}  \\
		\cline{3-8}
		& & $T1$  & $T2$ & $T3$ & $T1$  & $T2$ & $T3$  \\
		\midrule
		2021 & Jan--Mar & $10.5\%$    &  $8.54\%$  & $7.41\%$  & $7.78\%$  & $6.27\%$  & $2.87\%$   \\
		\midrule
		2018-- & Rare & $5.48\%$  &  $5.71\%$  & $5.60\%$ & $4.21\%$  & $3.89\%$  & $3.58\%$  \\
		2019 & Message & $4.12\%$    &  $4.88\%$  & $7.12\%$  & $3.07\%$  & $3.74\%$  & $6.05\%$  \\
		 & General & $8.25\%$ &  $7.88\%$  & $7.30\%$  & $5.07\%$  & $5.03\%$  & $6.55\%$    \\
		\bottomrule            
	\end{tabular}
\end{table}
\noindent Table~\ref{tab:teachers} shows the WERR of two student models ($rnnt\_90m$ and $rnnt\_60m$) when trained using the machine transcripts generated from the three teacher models. We observe that both the students are similar in terms of performance. On an average, for $rnnt\_90m$, $T1$ based training is better than $T2$ and $T3$, with $4.66\%$ and $3.25\%$ WERR, respectively. For $rnnt\_60m$, $T1$ is better than $T2$ and $T3$ with $5.96$ and $2.18\%$ relative WERR improvement respectively. The improvements are larger than in Table \ref{tab:replay-vs-no} as these experiments were done with 3 months of data using a weaker baseline. In fact, both the students have same order of performance as the teachers, that is $T1>T2>T3$ even after training in IL on new data. More important, the magnitude of improvement in student (true for both the student models) training is not of same scale as the difference in teachers. For example, Conformer based teacher ($T1$) is better than $T3$ by $7.96\%$, whereas $rnnt\_90m$ student trained with Conformer transcripts ($T1$) is $3.25\%$ better than the one trained with $T3$ transcripts. This suggests that better teacher models result in improving the student performance but the difference (same student trained with different teacher models) is narrower. In other words, a significantly better teacher model can have a limited impact in improving students models in ILASR.

\section{Related Work}
\label{sec:related-work}
\textbf{SGD gradients mini and large:} Stochastic gradient descent (SGD) drives the training of neural nets with mini batches. Large mini batches~\cite{goyal2017accurate, hoffer2017train, you2017scaling, smith2018dont} reduce the number of updates with a large step size. Simply increasing the batch size reduces the test accuracy~\cite{keskar2016large} as the gradients get integrated. Test set accuracy can be improved with large batches that are proportional to the learning rate. This simple linear scaling is inefficient, which necessitates a warm-up phase~\cite{goyal2017accurate}. Instead of decaying the learning rate, increasing the batch size during training~\cite{smith2018dont} helps to reduce the communication steps to update the model and improves the test accuracy. Federated averaging~\cite{mcmahan2017communication} (FedAvg) follows a similar strategy of synchronously updating the gradients. Thus, the centralized model simply aggregates the updates from various clients. Therefore, we apply these large synchronous batch updates (as in~\cite{smith2018dont}) to the model in federated settings (similar designs were proposed in~\cite{bonawitz2019towards}) both in federated SGD and averaging algorithms. Considering the negative effects of large-batch training on test accuracy in literature~\cite{KaiICASSP,keskar2016large,LI:KDD,shallue2018measuring,mccandlish2018empirical,golmant2018computational,masters2018revisiting}, a post-local SGD was proposed~\cite{Lin2020DontUL}, inspired from the FedAvg, where they adopt a warm-up~\cite{goyal2017accurate} based mini-batch SGD training for initial training before launching FedAvg. Similarly, distributed SGD for speech~\cite{strom2015scalable} and large scale training with million hours of speech~\cite{parthasarathi2019lessons} have helped accelerate the production for ASR models.

\textbf{Semi-supervised Learning in ASR:} The semi-supervised learning described in~\cite{karita2018semi, karita2019semi} employed auto-encoders to extract speech and text features from unpaired text and speech data. Semi-supervised ASR with filter-bank features~\cite{ling2020deep} use deep contextualized acoustic representations with small amounts of labeled data. The weak distillation of audio-text pairs resulted from the unsupervised techniques in~\cite{li2019semi} helped in improving the end to end ASR. The semi-supervised approaches in~\cite{weninger2020semi} combined the data augmentation through spectral augment~\cite{park2019specaugment} and consistency regularization to improve the performance. Dropout offer the power of ensembles, the semi-supervised dropout attempts in~\cite{dey2019exploiting} improved the pseudo label accuracy and model performance in ASR. Recently, the work in~\cite{xiao2021contrastive} employed contrastive semi-supervised learning with pseudo-labeling in transcribing the video content. In this paper, we use the pseudo-labels generated from a teacher model in federated setting with large batch sizes.

\textbf{Unsupervised Learning in ASR:} A related area is the training of representation, foundation, or upstream models from scratch using large volumes of unlabelled data. This model can then be fine-tuned for downstream use cases such as ASR, speaker recognition, among others. This paradigm is contrasted with the use case of incremental updates to a pre-trained ASR model presented in this work. A comprehensive survey of such methods for speech representation learning are in \cite{mohamed2022self}. The upstream model is trained with a \emph{pretext task} such as a generative approach to predict or reconstruct the input given a limited view (eg past data, masking) such as autoregressive predictive coding \cite{chung2020generative}. In a contrastive approach, a representation is learned that is close to a positive sample and further away from negative samples; wav2vec 2.0 \cite{baevski2020wav2vec} is an exemplar where the representation is trained to be close to a quantized target vector. Finally, in predictive approaches \cite{hsu2021hubert,chen2021wavlm,baevski2022data2vec}, the pretext task is to predict for masked input timeframes, a distribution over a discrete vocabulary such as clustered log-mel features. ASR models pre-trained using these techniques can be updated using ILASR. 

\section{Conclusions}
\label{sec:conclusions}
We proposed the ILASR framework for privacy preserving incremental learning of end-to-end automatic speech recognition systems. ILASR is a big step forward for production level ASR systems, especially for automatic incremental updates of these systems. In this study of near-real time training with ILASR, we learned that even the converged production level ASR models: 1) can be improved significantly in an incremental fashion with 3\% general improvements that can go up to 20\% on test sets with new words or phrases; 2) training with large batches arising as a result of communication constraints does not result in degradation; 3) memory replay training is effective at mitigating catastrophic forgetting on older test sets; 4) there is no significant impact of chronological versus random processing of data in IL for speech recognition over a period of six months; and finally; 5) having a significant improvement in teacher models used to generate machine transcripts does not translate to the same scale of improvements in students. 

In the future, we will explore the utility of noisy students for iterative self-learning instead of relying on teacher models in ILASR. Real-time resource-constrained on-device speech recognition is still a hard challenge. Here, we plan to further explore different directions such as finding the best hyper parameters~\cite{khodak2020weight}, controlling leaky gradients~\cite{zhu2019deep}, stopping gradient inversion and data leakage attacks~\cite{triastcyn2020federated}, personalizing ASR depending on the device context, and using smaller teacher models or self-labelling that can be run on device. Approximate gradient computation techniques may be required with severe compute resource limitations. Further, exploring methods of integrating weak supervision information from inferred or explicit user feedback from a session of interactions as well as externally updated language models are avenues of further research.
\section*{Acknowledgments}
We thank Kishore Nandury, Fred Weber, and Anand Mohan for discussions related to production ASR and utterance selection heuristics. Valentin Mendelev assisted with delta testset construction to measure the impact of IL on new data. We thank Bach Bui, Ehry MacRostie, Chul Lee, Nikko Strom,  and Shehzad Mevawalla for helpful discussions, review and support. We are indebted to the Alexa Speech Recognition group for comments, dataset construction and training infrastructure development. 


\end{document}